\documentclass[10pt,conference,a4paper]{IEEEtran}

\usepackage{graphics}

\usepackage{epsfig}
\usepackage{amsmath}

\usepackage{makeidx}
\usepackage{graphicx}
\usepackage{amsmath,amssymb} 
\usepackage{color}
\usepackage{algorithm}
\usepackage[noend]{algpseudocode}

\ifCLASSINFOpdf
\else
\fi

\begin{document}

\title{ Layer-wise training of deep networks \\ using kernel similarity }

%

\author{Mandar Kulkarni \hspace{0.5cm} Shirish Karande \\ TCS Innovation Labs, Pune, India \\ 
Email: (mandar.kulkarni3, shirish.karande)@tcs.com
}


\maketitle

\begin{abstract}
Deep learning has shown promising results in many machine learning applications. 
The hierarchical feature representation built by deep networks enable compact and precise encoding of the data. 
A kernel analysis of the trained deep networks demonstrated that with deeper layers, more simple and more accurate data representations are obtained.
In this paper, we propose an approach for layer-wise training of a deep network for the supervised classification task.
A transformation matrix of each layer is obtained by solving an optimization aimed at a better representation where a subsequent layer builds its representation on the top of the features produced by a previous layer. We compared the performance of our approach with a DNN trained using back-propagation which has same architecture as ours. 
Experimental results on the real image datasets demonstrate efficacy of our approach.
We also performed kernel analysis of layer representations to validate the claim of better feature encoding.

		
	\end{abstract}

	\section{Introduction}
	\label{sec:introduction}

Deep learning has shown promising results in many machine learning applications. In computer vision, it has been successfully applied to problems such as object detection \cite{ref2}, English character recognition \cite{ref3}. It also showed promising results for speech data where it has been applied for speech recognition \cite{hinton2012deep} and spoken keyword spotting \cite{ref4}.
It has been studied that the effectiveness of deep networks lies in the layered representation \cite{hinton2007learning}. 
A deep learning architecture automatically learns the hierarchy of feature representations where complex features are built on the top of the simple encodings. Higher layers construct more abstract representation of the input data enabling well-generalizing representations.

Recently, there has been an attempt to analyze the layered representation built by a trained deep network for the classification task. Montavon et. al.
\cite{montavon2011kernel} performed the kernel analysis of the data representations at each layer of the network trained for a supervised classification task. 
The representation is analyzed with the kernel Principal Component Analysis (kPCA) \cite{bulthoffnonlinear} using a RBF kernel. At each layer, a kernel matrix is computed on the feature representation of the training data. The kernel space representation of the features is computed by projecting data onto leading $v$ Eigen vectors of the kernel matrix. 
The dimensionality $v$ controls the simplicity of the model. A smaller $v$ corresponds to a simpler model while large value of $v$ indicates relatively complex model.
A linear classifier is then trained to this low rank representation to compute the training error rates. 
Interestingly, it has been observed that as deeper and deeper kernels are built, simpler and more accurate representations of the learning problem are obtained \cite{montavon2011kernel}.
In other words, as representation evolves layer by layer, more energy gets concentrated in fewer and fewer leading Eigen vectors. The trend is observed in representations built by a Multi-Layer Perceptron (MLP) as well as Convolutional Neural Networks (CNN).


For a classification task, MLPs compute the complex mapping from input to output using the multiple fully connected layers.
The weights in the network are updated using the back-propagation algorithm.
Due to the large number of trainable parameters, MLPs often suffer from issues such as slow training process, over-fitting and large computation resources.

In this paper, we propose a novel layer-wise training approach for deep networks aimed at supervised classification. 
For each layer, we aim to find a transformation which enables a better representation of the data. 
A kernel matrix is defined through the transformation matrix of the layer.
We setup an optimization problem to compute a desired layer transformation. 
Our optimization attempts to render a kernel increasingly more similar to the ideal kernel matrix. In an ideal kernel matrix, data points from a same class have kernel value equal to one while data points from different classes have zero similarity. 

To evaluate a goodness of representations obtained layer after layer, 
we compared our results with a standard DNN trained with back-propagation which has exactly same architecture as ours.
We observe that classification accuracies obtained with our approach are comparable to DNN accuracies.  
In addition to this, we also perform a kernel PCA analysis similar to \cite{montavon2011kernel},\cite{montavon2010layer}. Experimental evaluation demonstrate that our layer-wise training method indeed produce better representations layer by layer. 


We performed extensive experiments on the publicly available image recognition (classification) datasets. Experimental results demonstrate that, instead of fixing an architecture a priori, there is a possibility of selectively adding layers to the network with guaranteed improved performance.
It hints towards the feasibility of a modular approach for deep learning.

\section{Related works}
Layer wise training of deep networks in an unsupervised manner is a well known technique in the literature. 
Bengio et. al. \cite{bengio2007greedy} proposed greedy layer-wise unsupervised
training strategy for training a multi-layer deep network. It was demonstrated that the unsupervised training 
leads to a good generalization performance by appropriately initializing weights in a
region near a good local minimum which provides feature representations
that are high-level abstractions of the input.
Larochelle et. al. \cite{larochelle2009exploring} confirmed the hypothesis that the greedy layer-wise unsupervised training
strategy indeed helps the optimization by initializing weights near a good local minimum. Also, it implicitly acts as a regularization that brings better generalization and produces good internal data representations. 

Autoencoders play an important role in unsupervised learning for deep architectures.
Baldi et. al. \cite{baldi2012autoencoders} thoroughly studied the linear and non-linear auto encoders in the aspects such as their learning complexity, their horizontal and vertical composability in deep architectures and their fundamental connections to clustering, Hebbian learning, and information theory.
Using a different aspect, Xu et.al. \cite{xu1999training} demonstrated 
how information potential can be used to train a
MLP (multilayer perceptron) layer-by-layer. It was demonstrated that the hidden layer of a MLP serves as an
information filter which tries to best represent the
desired output in that layer in the statistical sense of
mutual information.

There has been efforts to analyze the hierarchical feature representation built by deep networks.
Montavon et. al. \cite{montavon2011kernel},\cite{montavon2010layer} performed a kernel analysis of a deep networks such as Convolution Neural Network (CNN), Multi-layer Perceptron (MLP) and  Pretrained Multi-layer Perceptron (PMLP) trained for a supervised multi-class classification task. 
Feature representation at each layer is analyzed for its complexity and accuracy with kernel PCA using a Gaussian kernel. It was observed that all deep networks create increasingly better representations
of the learning problem layer by layer. Also, the structure (or type) of a deep network has an impact on the way representations are built. It was demonstrated that a layer wise representation built by CNN differs from that of MLP. Also, MLP starts the discrimination from a first layer itself while CNN postpones the discrimination task to later layers.

\section{Methodology}
In this section, we describe an approach used for training MLP layer by layer. We use image datasets to validate the proposed approach. However, our training procedure is quite generic in nature and should work with other datasets as well. 

\subsection{Data pre-processing}
We perform pre-processing on the input features prior to using them for training a layer. 
Each feature vector is normalized independently by subtracting its mean value and dividing by its norm. 
We observe that this process is required to avoid the saturation of neurons due to non-linarity.

\subsection{Layer training}
\begin{figure} [!h]
\centering
\begin{tabular}{c}
\includegraphics[height = 120pt,width = 220pt]{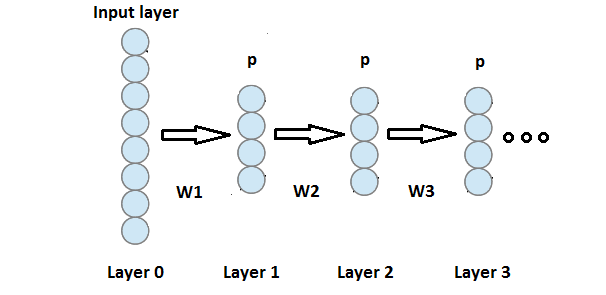}\\
\end{tabular}
\caption{\label{fig:sv} Our network architecture. p indicates the layer dimension while W indicates the transformation matrix associated with a layer.}
\end{figure}

A MLP build a hierarchy of feature representation where a subsequent layer build a representation on the top of the features computed by previous layers. 
An input to our first layer is the set of $n$ labeled images [$(t_1 , l_1), ..., (t_n , l_n)$] where $t$ denotes a vectorized training image and $l$ denotes the corresponding label. If $M$ and $N$ denotes the dimension of an input image, $t \in R^{d}$ where for a color image, $d = MN \times 3$ while for a grayscale image, $d = MN$. Input data is projected onto a $p$ dimensional space where $p$ indicates the layer dimension.
Fig. \ref{fig:sv} shows our network architecture. 

An operation at the $k^{th}$ layer of a MLP can be represented as follows
\begin{eqnarray}\label{eq:m4}
X_k = \tanh (D_{k-1} W_k ) \hspace{0.3cm} D_{k-1} \in R^{n \times d}  \hspace{0.2cm} W_k \in R^{d \times p}
\end{eqnarray}

\noindent where $D_{k-1}$ denotes the training data matrix, $W_k$ denotes the weight matrix for the $k^{th}$ layer. Feature vectors of the training data matrix $D_{k-1}$ are appended with one to account for the bias term.

Since the first layer directly interacts with the input data, $D_0$ indicates the preprocessed input image data.
At the start, our attempt is to calculate an appropriate transformation matrix $W_1$ such that the feature representation at the output of the first layer is more favorable for the classification task.
Since we are performing the multi-class classification task, 
in the ideal kernel matrix, points from the same class should depict the maximum similarity while points from different classes should have least similarity.
Therefore,
we specify a target kernel function as follows  
\[
    T(i , j) = 
\begin{cases}
    1,& \text{if } l_i = l_j \\
    0,              & \text{otherwise}
\end{cases}
\]
\noindent where $l_i$ and $l_j$ denotes the labels of the $i^{th}$ and $j^{th}$ training points, respectively. 

The kernel function to be aligned with a target function is defined through the layer transformation matrix $W_1$ as follows. 
Let $K$ denotes the kernel matrix computed on the output of the layer. We use a Gaussian kernel defined as follows 
\begin{eqnarray}\label{eq:m1}
K(i , j) = \exp \frac{-|| x_i - x_j ||^2}{2 \sigma^2} \hspace{0.2cm} 1 \leq i \leq n \hspace{0.2cm} 1 \leq j \leq n  
\end{eqnarray}
\noindent where $x_i$ and $x_j$ denotes the feature representation for $i^{th}$ and $j^{th}$ data sample at the output of the first layer. 
Each feature vector is made unit norm as follows
\begin{eqnarray}\label{eq:a3}
x_i =  x_i / || x_i ||_2
\end{eqnarray}

\noindent The squared Euclidean norm between any two vectors can be expanded as follows
\begin{eqnarray}\label{eq:a5}
|| x_i - x_j ||_2^2 = ||x_i||_2^2 + ||x_j||_2^2 - 2 x_i^{T}x_j 
\end{eqnarray}

\noindent Using Eq. \ref{eq:m1},\ref{eq:a3} and \ref{eq:a5}, a similarity between two vectors can be obtained as 
\begin{eqnarray}\label{eq:a1}
K(i , j) = \exp \frac{-2 (1 - x_i^{T}x_j )}{2\sigma^2}  
\end{eqnarray}

\noindent A kernel matrix $K$ can thus be calculated as 
\begin{eqnarray}\label{eq:m2}
K = \exp \frac{- (1 - X^{T}X) }{\sigma^2}  
\end{eqnarray}

\noindent Now, we solve a following optimization problem 
\begin{eqnarray}\label{eq:m6}
\min_{W_1}  \hspace{0.2cm}  \frac{1}{n^2}||K - T||_F^2  +  \lambda ||W_1||_2^2
\end{eqnarray}
\noindent Here, $||W_1||_2^2$ denotes the L2 regularization term where $\lambda$ controls the degree of regularization. 

The optimization is solved using a gradient descent where    
the weight matrix $W_1$ is initialized with values from a normal distribution with zero mean and unit standard deviation $\mathcal{N}(0 , 1)$. 
The optimal solution obtained is used as the transformation matrix for the first layer.
Input data is transformed as described in Eq. \ref{eq:m4} which provides the feature representation at the first layer. 

The subsequent layers are trained using the same procedure as described above where the feature representation provided by the previous layer is considered as an input to the current layer. Data pre-processing is done prior to the calculation of each layer transformation.
The training procedure for a layer is described in Algorithm Block \ref{euclid}.

\begin{figure*} [!t]
\centering
\begin{tabular}{c c c}
\includegraphics[height = 120pt,width = 120pt]{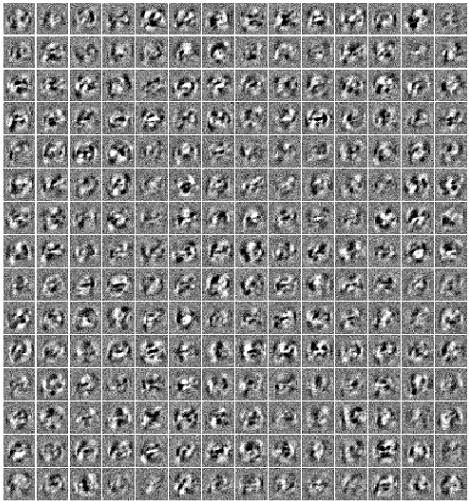}&
\includegraphics[height = 80pt,width = 250pt]{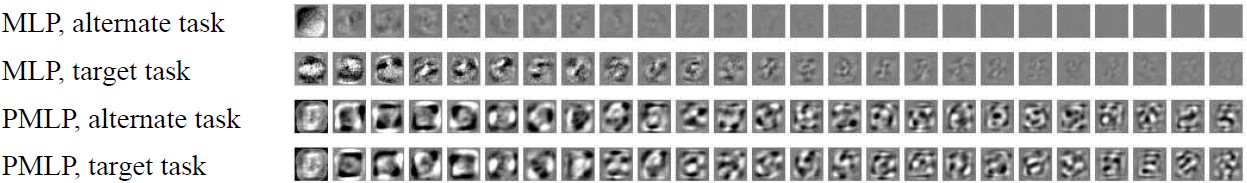}&
\includegraphics[height = 120pt,width = 120pt]{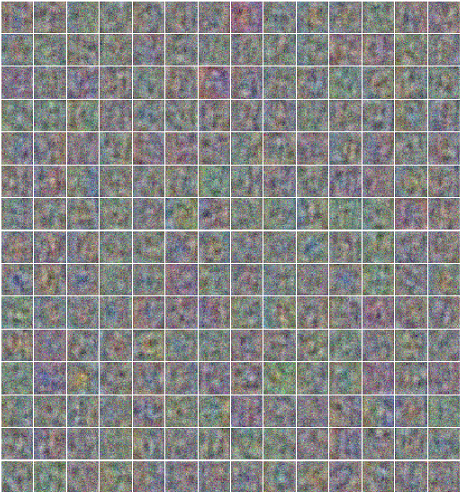}\\(a)&(b)&(c)\\

\end{tabular}
\caption{\label{fig:mnist2} Features learned at the first layer of network for MNIST and CIFAR-10 dataset. (a) first layer 210 features of MNIST dataset using our approach, (b) first layer features for MLP and Pre-trained MLP (PMLP) trained on target and alternate tasks as explained in \cite{montavon2010layer}. Note that our feature visually look similar to PMPL first layer features (c) first layer 210 features of CIFAR-10 dataset with our approach. }
\end{figure*}			

\begin{algorithm}
\caption{Training procedure for a $k^{th}$ layer of a MLP given previously trained ($k - 1$) layers}\label{euclid}
\begin{algorithmic}[]
\Procedure{Training a $k^{th}$ layer}{}

\State $\textit{$W_k$} \gets \textit{Random initilization}$ ( $10^{-3} \times\hspace{0.1cm} \mathcal{N}(0 , 1)$)

\State \textit{Until convergence}:
\State $ X_k = \tanh(D_{k-1} W_k)$
\State \textit{Make feature representations zero mean and unit norm}
\State $ K = \exp \frac{- (1 - X_k^{T} X_k)}{\sigma ^ 2}$
\State  \textit{cost} $  = \frac{1}{n^2}||K - T||_F^2 + \lambda ||W_k||_2^2 $
\State $ gW = \frac{d}{d W_{ij}}$ \textit{cost}
\State $ W_k = W_k - \mu * gW $ 

\EndProcedure
\end{algorithmic}
\end{algorithm}

Since the transformation which minimizes the error between current kernel and an ideal kernel is sought, the kernel distance between points from the same class decrease while distance between points from different classes increase. Therefore, the transformation matrix computed at each layer projects data onto the space which is more favorable for the classification. 

To perform the classification, any off-the-shelf non-linear classifier can be trained on the features at the output of a layer. To speed-up the computations at the test time, we train a fully connected layer (of dimension 100) and a softmax unit on these features. Note that the fully connected layer is trained directly on the features learned by a layer and training does NOT involve updation of previously calculated features. We will denote this fully connected layer as FC100.

\section{Network evaluation}
To validate the efficacy of our approach, we designed two evaluation procedures as described below.

\subsection{Kernel PCA analysis procedure}
Let $X_k$ denotes the feature representation of training data at $k^{th}$ layer.
Let [$(x_{k1}, l_1) , (x_{k2}, l_2), ..., (x_{kn}, l_n)$] denote individual feature vectors and corresponding labels of the training data. $x_{li}$ denotes the feature representation of $i^{th}$ data point and $l_i$ denotes the training label for the data point (image).
The feature representation at each layer is analyzed similar to the procedure as described in \cite{montavon2010layer}.

The kernel matrix $K$ is computed on the feature set $X_k$ as described in Eq. \ref{eq:m1}.
Eigen value decomposition of the kernel matrix is obtained as
\begin{eqnarray}\label{eq:m3}
K = U \Lambda U^{T} 
\end{eqnarray}
\noindent where Eigenvectors (columns of $U$) have unit length and eigenvalues (diagonal entries in $\Lambda$) are sorted by decreasing magnitude.

Let $U_d$ and $\Lambda_d$ denotes the $d$ dimensional approximation of Eigen decomposition. $U_d$ denotes the kernel representation of the data points. A logistic regression classifier is trained on this low dimension representation. 
The optimal parameters $\beta^*$ of the logistic regression are learned as follows

\begin{eqnarray}\label{eq:m3}
\beta^* =  \arg \min_{\beta} \prod_{i=1}^{n} \mbox{softmax}( [U_d \beta]_i )_{l_i}   
\end{eqnarray}
\noindent where $\beta^*, Y \in R^{n \times c}$ where $c$ denotes number of classes. The $\mbox{softmax}$ is defined as follows
\begin{eqnarray}\label{eq:m3}
\mbox{softmax}(z) = \frac{e^{z}}{\sum_j{e^{z^{j}}}}   
\end{eqnarray}

The class for the $i^{th}$ test points is then estimated as 
\begin{eqnarray}\label{eq:m3}
 \widehat{l_i} = \arg \max([U_d \beta^*]_i)  
\end{eqnarray}

%

The classification error for the $d$ dimensional representation is then calculated as follows
\begin{eqnarray}\label{eq:m3}
e(d) =  1 - \frac{1}{n}\sum 1_{\widehat{l} == l }
\end{eqnarray}
\noindent where $l$ and $\widehat{l}$ denote ground truth and estimated labels respectively.


\subsection{Comparison with DNN}
To evaluate the classification performance of our approach, we compared our results with a DNN.
To have a valid comparison, we train a DNN keeping exactly the same architecture as ours e.g. if we only train a single layer, the DNN will have a fully connected layer of dimension $p$, FC100 layer and a softmax unit. 
We also included the L2 regularization while training the DNN.
The whole network is trained in the supervised fashion using back-propagation. In the experimental results, we demonstrate the classification accuracy comparison on multiple datasets with varying sizes of training samples.

\begin{figure} [!h]
\centering
\begin{tabular}{c c}
\includegraphics[height = 100pt,width = 120pt]{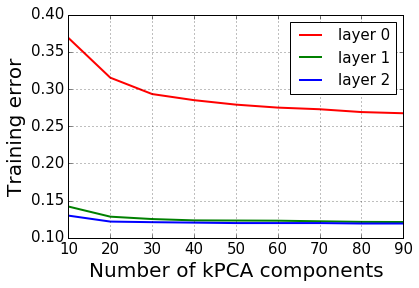}&
\includegraphics[height = 100pt,width = 120pt]{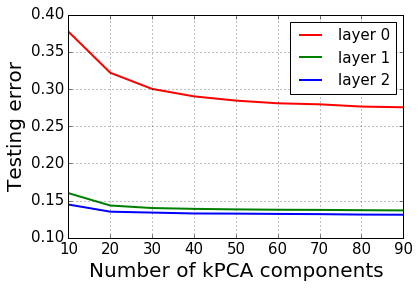}\\(a)&(b)\\

\end{tabular}
\caption{\label{fig:mnist1} Kernel PCA analysis for MNIST dataset. (a) training error for different kPCA components, (b) testing error for different kPCA components. Layer 0 indicates the input (image) feature layer.}
\end{figure}		
						
\section{Experimental results}
To validate the efficacy of our method, we performed experiments with real world image datasets such as MNIST handwritten digit dataset \cite{ref3}, CIFAR - 10 image dataset \cite{krizhevsky2009learning}.
	
\subsection{MNIST handwritten digit dataset}
MNIST dataset consist of grayscale images of handwritten digits. Each image is of dimension $28 \times 28$.  
The dataset consist of 60k training images and 10k testing images.	

In order to perform kernel PCA analysis on layer representations, we sequentially trained two layers, on the 5k training samples of the MNIST dataset. The layer dimension $p$ is set to 784 for each layer. $\sigma$ value was set to 1 for all the experiments.
Fig. \ref{fig:mnist1} shows the training and testing errors obtained at three layers through the kernel analysis.
From Fig. \ref{fig:mnist1}(a), it can be seen that, as deeper representations are built, more energy is getting concentrated in fewer Eigen vectors with better training accuracy (lower training error).
Fig. \ref{fig:mnist1}(b) shows the graph of the testing error. It can be seen that, our representation agrees with the kernel analysis results reported in \cite{montavon2010layer}. 

Fig. \ref{fig:mnist2}(a) shows filters obtained at the first layer. Fig. \ref{fig:mnist2}(b) shows the filters obtained in the first layer of MLP and Pre-trained MLP (PMLP) when an architecture of four layers is trained \cite{montavon2010layer}.  
Note that, unlike the MLP filters displayed in Fig. \ref{fig:mnist2}(b), our approach produce rich features.  
Interestingly, our first layer features visually look similar to those of PMLP. 
 		
\begin{figure} [!h]
\centering
\begin{tabular}{c c}
\includegraphics[height = 100pt,width = 120pt]{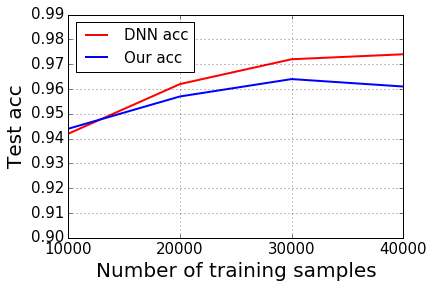}&
\includegraphics[height = 100pt,width = 120pt]{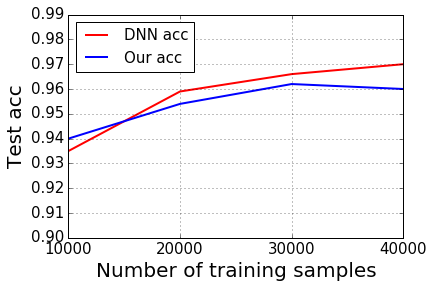}\\(a)&(b)\\

\end{tabular}
\caption{\label{fig:mnist3} Test accuracy comparison with DNN for varying sizes of training set. (a) a single layer trained with our approach, (b) two layers trained with our approach. The performance of the proposed method is comparable with DNN.  }
\end{figure}

We also compared the classification performance of our approach with the DNN. Initially, we trained a layer using the proposed optimization and 
calculated the classification test accuracy by subsequently training a FC100 and softmax layer. 
We then trained a DNN with fully connected layer of dimension 784, FC100 and a softmax layer. 
The FC100 $+$ softmax layer in our approach and a DNN used for comparison are trained until convergence.  
Fig. \ref{fig:mnist3}(a) shows the test accuracy comparison for varying sizes of training set.
Subsequently, we trained an another layer on top of the features exacted from first layer. We also trained a DNN  with same architecture. Fig. \ref{fig:mnist3}(b) shows the comparison plot.
From these plots, it can observed that our approach indeed computes discriminative features which are suitable for classification.
This underlines the efficacy of our layer-wise training and it effectiveness for object recognition.
	
	\subsection{CIFAR-10}	
	
\begin{figure} [!h]
\centering
\begin{tabular}{c c}
\includegraphics[height = 100pt,width = 120pt]{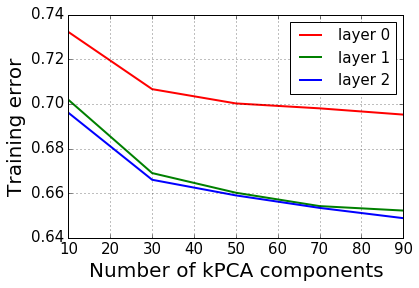}&
\includegraphics[height = 100pt,width = 120pt]{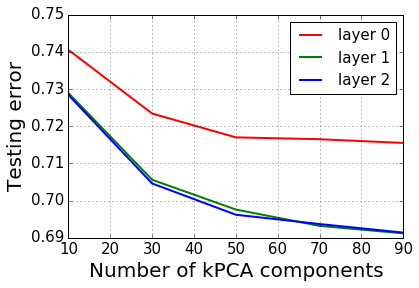}\\(a)&(b)\\

\end{tabular}
\caption{\label{fig:cifar1} Kernel PCA analysis for CIFAR-10 dataset. (a) training error for different kPCA components, (b) testing error for different kPCA components. Layer 0 indicates the input (image) feature layer.}
\end{figure}

CIFAR- 10 dataset consist of color images of dimension $32 \times 32$ corresponding to 10 object classes. 
It contains 50K training samples and 10k testing samples.
We performed an experiment similar to the MNIST dataset where layers are learned sequentially. The layer dimension was set to 1500  and $\sigma$ value was set to 1 for the experimentation. 
Using the proposed optimization, we trained two layers, one after the other.
The kernel PCA analysis was performed on the layered representation built with 5k training samples. Fig. \ref{fig:cifar1} shows the training and testing error pattern with different number of kernel PCA components. Note that, here as well, simpler and more accurate representations of the input are obtained using our approach. Fig. \ref{fig:mnist2}(c) shows the first layer filters. Note that, our approach learns rich features which enable better discrimination.   
		
	
\begin{figure} [!h]
\centering
\begin{tabular}{c c}
\includegraphics[height = 100pt,width = 120pt]{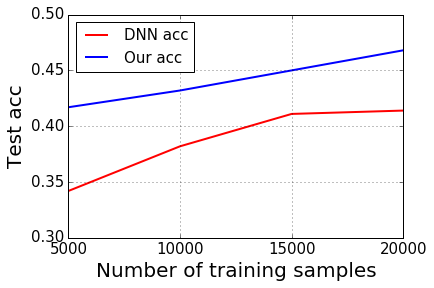}&
\includegraphics[height = 100pt,width = 120pt]{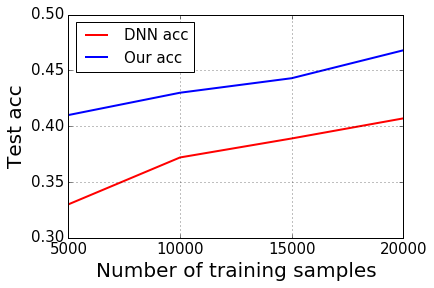}\\(a)&(b)\\

\end{tabular}
\caption{\label{fig:cifar2} Test accuracy comparison with DNN for varying sizes of training set. (a) a single layer trained with our approach, (b) two layers trained with our approach. The proposed method outperforms DNN in both the cases.  }
\end{figure}

Next, we compared the results with a DNN for varying sizes of the training set. Fig. \ref{fig:cifar2} shows the test accuracy comparison with a DNN trained in the supervised setup. Note that, even with less amount of training data, our approach outperforms the DNN. This underlines the effectiveness of layer-wise training where the effect of overfitting can be minimized since the number of parameters to be tuned at a time are small.

			%
			%
%
%
			%
					%

\section{Conclusion}
In this paper, we proposed a kernel similarity based optimization for layer-wise training of deep networks. The optimization attempts to compute a linear transformation followed by non-linearity which renders kernel at each layer increasing more similar to the ideal kernel. The kernel analysis of the layer-wise training demonstrate that with each layer, better representation of the input data are obtained. 
Since we train a layer at a time, the number of parameters to be tuned are relatively small. Hence, the effect of overfitting is minimized which is clearly evident from the result on CIFAR-10 dataset.
We also compared our classification accuracies with the DNN trained in the supervised fashion. Experimental results show that our approach produce comparable results with that of DNN.

\bibliographystyle{splncs03}
\bibliography{papers_bib}

\IEEEpeerreviewmaketitle

\end{document}